# Deviant Learning Algorithm: Learning Sparse Mismatch Representations through Time and Space

*E.N. Osegi
Department of Information and Communication Technology
National Open University of Nigeria
Lagos State, Nigeria

System Analytics Laboratories (SAL)
Sure-GP Ltd
Port-Harcourt
Rivers State, Nigeria.
E-mail: nd.osegi@sure-gp.com

V.I.E Anireh
Department of Computer Science
Rivers State University of Science and Technology
Port-Harcourt
Rivers State, Nigeria.
E-mail: anireh.ike@ust.edu.ng

**Abstract**

Predictive coding (PDC) has recently attracted attention in the neuroscience and computing community as a candidate unifying paradigm for neuronal studies and artificial neural network implementations particularly targeted at unsupervised learning systems. The Mismatch Negativity (MMN) has also recently been studied in relation to PDC and found to be a useful ingredient in neural predictive coding systems. Backed by the behavior of living organisms, such networks are particularly useful in forming spatio-temporal transitions and invariant representations of the input world. However, most neural systems still do not account for large number of synapses even though this has been shown by a few machine learning researchers as an effective and very important component of any neural system if such a system is to behave properly. Our major point here is that PDC systems with the MMN effect in addition to a large number of synapses can greatly improve any neural learning system's performance and ability to make decisions in the machine world. In this paper, we propose a novel computational intelligence algorithm – the Deviant Learning Algorithm, inspired by these key ideas and functional properties of recent brain-cognitive discoveries and theories. We then compare the DLA's performance with the Hierarchical Temporal Memory (HTM) based on cortical learning algorithms for a set of classification tasks.

*Corresponding Author

1. **Introduction**

It is well known fact that brain-like learning in machine oriented systems have significant benefits in real world and synthetic data problems. However, the most suitable brain-cognitive approach to the data learning problem still remains a primary problem and have troubled researchers world-wide. While some algorithms do very well in certain tasks, they perform poorly in others confirming the "No-Free Lunch Theorem" (NFLT). Nevertheless, competitive machine learning systems that employ brain-like algorithms have evolved over time and present a promising alternative to a wide variety of machine learning algorithms. Two algorithms stand out as major candidates for future ML systems. The Hierarchical Temporal Memory based on the Cortical Learning Algorithm (HTM-CLA) proposed in [1] and the Long Short-Term Memory (LSTM) in [2] have shown promising results in recent times; however, HTM-CLA seems to model closely the intelligent behavior of cortical micro-circuits. Currently, the HTM-CLA is on its second phase of development with the primary purpose of synthesizing a biologically plausible algorithm [3]. In this paper, we introduce yet another model of brain-like learning. This model is inspired by the Mismatch Negativity (MMN) effect elicited in real humans. HTM-CLAs have the unique ability to predict its own activation and they offer a generative procedure for building cellular data; Cellular data is a very important component of hierarchical memory-based learning systems as they allow the unique predictive capabilities found in most cortical generative models.

HTM-CLA do not account for MMN effect. Specifically they do not use our deviant (Real Absolute Deviation (RAD)) modeling approach. As described in the Numenta white paper [4], post-predictive ability with slowness through time is a very essential component of machine learning systems with intelligent abilities. Our model possess the unequivocal and unique ability to enforce slowness through time; it incorporates some of the ideas of predictive coding based on the MMN effect as described in [5-6] to evolve a deviant model of the world. It also encourages the use of the overlap concept as in HTM-CLA with good post-predictive abilities. In line with the six principles of biological based cortical cognition computational models [7], our model may be described as biologically plausible.

Our primary goal here is to develop a model algorithm that builds on the strengths of

CLA. Such a model should also account for the MMN effect as this can prove useful in a wide range of computational problems.

This paper is organized as follows:

In section 2 we briefly review related works that inspired this paper. Brain-like theories and ideas for deeper understanding of this paper is presented in Section 3 while Section 4 describes the Deviant Learning algorithm (DLA) based on the MMN effect. Next, we present and describe comparative results on a set of classification tasks (using the IRIS, HEART and Word-Similarity benchmark datasets) between the DLA and the HTM-CLA; in addition, we study the behavior of the DLA for varying learning extents with a discussion of results in Section 6. Finally, we give our conclusions and contributions to knowledge in Sections 7 and 8 respectively.

## 2. Related Works

Over the years, brain-like learning in artificial neural networks (ANNs) have been described by many artificial intelligence (AI) researchers and various models proposed as the quest for a better machine intelligence close to human reasoning continues worldwide. More importantly, several AI approaches have recently captured the interest of the artificial intelligence community some of which include slow feature learning in [8], the deep generative models in [9] and deep recurrent models in [10], history compression models with a focus on unexpected inputs during learning [11], the spatio-temporal hierarchical representations in [1] including the concept of sparse distributed representations as in [12]. Also worth mentioning is the spiking neuron familiarity and frequency models in [13], predictive coding model in [14], extreme learning machines [15] and neuro-computational models based on the Mismatch Negativity effect [6]. While all these models have one thing in common - they all employ some form of connecting link for processing sensory information, they still fall short of real human brain requirements. One interesting feature to note is the inability of most of these models to account for large number of synapses in the neocortex, the seat of intelligence in the brain - see for instance the argument in [16]. Nevertheless, arguments and counter-arguments have far reaching consequences and is still not sufficient to describe in algorithmic terms the primary requirements of a brain-like algorithm in software. Thus, the AI field is still at a loss at the most appropriate way to tackle the intelligent machine problem.

## 3. Machine Brain-like Effects and Novel Theories

Brain-like theories such as discussed extensively in [17] as a form of a memory-prediction framework, in [18] as deep generative connectionist learning and as recurrent learning networks with Long Short-Term Memory in [2, 19], have had profound effect in the solution approaches used in today's AI software systems. The HTM-CLA being a variant of the ideas introduced in [17] has particularly been applied to a variety of interesting real world tasks; for instance, they have been used for retina analysis with moderate performances reported in [20], sign language and gaze-gesture recognition [21-22] and weld flaw detection [23].

One interesting feature of some of these theories is the ability of machine-life algorithms to remember and forget - an analogue of the birth-death theory. We strongly believe in this behavior as artificial life machines try to respond to its world by the formation of new memories and the extinction of past memories. This re-creative process have strong genetic roots and is beyond the scope of this paper to describe every detail. In this section we discuss on some very important theoretical principles behind our deviant learning algorithm while inducing some novel ideologies in the process.

### 3.1 Mismatch Negativity (MMN) Effect

The Mismatch negativity effect (MMN) effect first discovered in the context of auditory stimulation in [24] is one possible neurobiological plausible attempt to model the behavior of real world neural tissue in a statistical way. It has also been studied in the context of visual modalities [25]. In particular the MMN effect seeks to validate the differential response in the brain. Some important MMN theories may be found in [6]. However, as discussed in [5], MMN theories are still debatable and not entirely conclusive up to this present moment. Some of the identified theories include [5]: the Change Detection Hypothesis (CDH), Adaptation Hypothesis (AH), Model Adjustment Hypothesis (MAH), Novelty Detection Hypothesis (NDH) and the Prediction Error Hypothesis (PEH). Notwithstanding these debates, MMN still provides a fundamental evidence that the brain learns a statistical structure of the world and is also capable of future predictions [5]. For instance the MMN effect can provide the enabling statistics for encoding actions given a sequence of observations; this functionality is a core feature of mirror neurons and permits the generalization of events [26]. Thus, the MMN effect can also be used to synthesize a new class of neurons we refer to here as "abstract neurons".

### 3.2 Reticular Formation and Synaptic Response

Responses to sensory impulses have historically been defined by the reticular forming units which serve as some sort of familiarity discrimination threshold detector with possible hypothalamic gating abilities [27]. This functionality also shares some resemblance to the notion of permanence introduced in [1]. Permanence have a depolarizing effect on the synapses leading to the state of "potential synapse" i.e. synapses with a high likelihood of being connected. Thus, when the synapse connects, the neuron or cell fires.

### 3.3 Incremental Learning and the Growing of Synapses

During post-predictive stage it is likely that the input exceeds the peak(s) of the generated units. In this situation it becomes necessary to grow more synapses in order to improve the learned observations in the neural ecosystem. This is achieved in the DLA using the parameter referred to as the learning extent. As an organism grows in age the learning extent should correspondingly increase. Thus, there is a gradual annihilation of previous primitive states to begin to learn more complex tasks. However, there is an extent to which this learning can go and the organism may lose some memory or may even become extinct (aging).

### 4. Deviant Learning Algorithm (DLA)

The Deviant Learning Algorithm (DLA) is approached using a systematic mathematical procedure. DLA in CPU-like structure is as shown in Figure 1. It is divided into two core phases - the pre-prediction and the post-prediction phases. During pre-predictions, invariant representations are learned by performing an input mismatch through a long generative list of well-defined standards. For the purposes of this study, the standards are assumed to be integer-valued representations of the input chain. Pseudo-code for the

DLA is provided in Appendix A.

### 4.1 Pre-Prediction Phase

Here we perform first and second-order mismatch operations with the assumptions that the inputs are single-dimensional (1-D) matrices or 1-D sequence of vectors. These operations are described in the following sub-sections.

### 4.1.1 Mismatch:

Suppose a permanence threshold range is defined as

$$\rho_o :$$
$$\rho_o \leftarrow \rho_1 .$$

Then, for every input deviant, we obtain a first-order (level-1) mismatch using a Real Absolute Deviation (RAD) as:

$$k_{dev(1)} = | I_i - I_{g(l)} |$$
$$= \begin{cases} 1 & k_{dev(1)} \leq \rho_1, \quad i=1,2,...,n, \ l \leq l_{ext}, \ n \in Z \\ 0 & otherwise \end{cases} \quad (1)$$

.where,

$I_{g(l)}$ .is a long list of generative integers limited by an explicit threshold, $l_{ext}$ and,

$I_i$ .is the source input for each exemplar at time step, t

The overlap (Deviant Overlap) is described as a summation over $k_{dev(1)}$, as frequent patterns of ones (binary 1's) - superimposed into a virtual store say $S_o$, conditioned on $k_{dev(1)}$ and is defined as:

$$S_{o_{(t,j)}} = S_{o_{(t,j)}} + 1$$

$$= \begin{cases} 1 & k_{dev(1)} \equiv 1 \\ 0 & otherwise \end{cases} \quad (2)$$

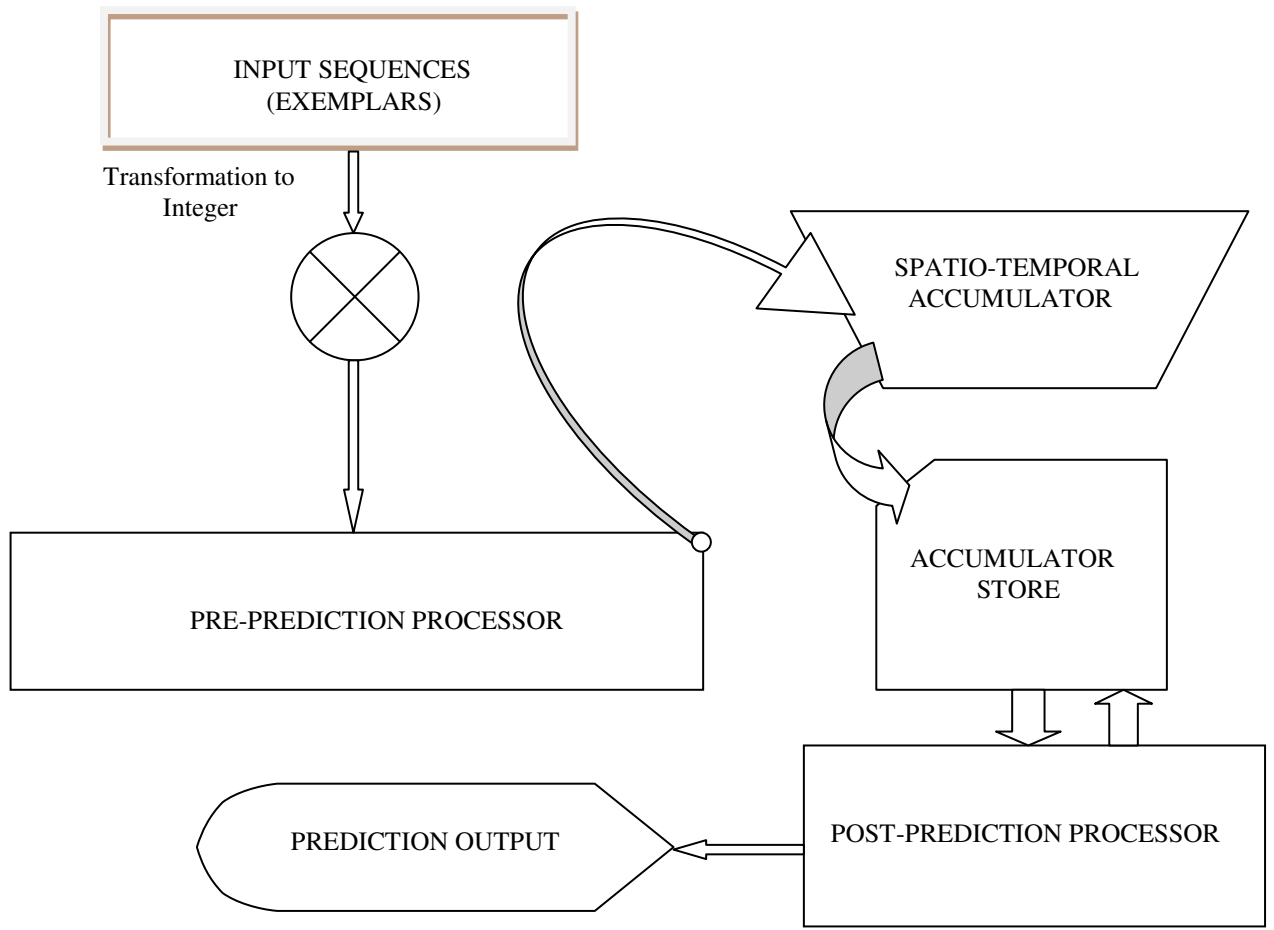

FIGURE 1: The DLA Systems Structure

From which we obtain:

$$S_{o_{(t)}} = \sum_{k_{dev(1)}=1}^{\vdots} S_o \qquad (3)$$

.and

$$S_{o_{(t)}}^{+} = \max(\max(S_{o_{(t)}})) \qquad (4)$$

### 4.1.3 Winner Integer Selection:

In order to identify the integer(s) responsible for the observation(s), $S_{o_{(t)}}^{+}$ is conditioned on a threshold say, $T_{h1}$ defined as a recurrent operation in the positive sense i.e. this time we select only those integers for which $S_{o_{(t)}}^{+}$ is greater than or equal to $T_{h1}$. The casual observers is described through the process of inhibition and by the recurrent operation:

$$C_{ao_n}^{(\cdots)} = C_{ao_{n-1}}$$
$$= \begin{cases} z & S_o^{+} \geq T_{h1} \\ 0 \end{cases} \qquad (5)$$

### 4.1.4 Learning:

We perform learning using a Hebbian-Hopfield update style conditioned on the maximum value of $S_{o_{(t)}}^{+}$ and a time-limit say $t_i$. We also additionally encourage diversity by adding some random noise to the permanence update procedure defined as:

$$\rho_1 = \rho_1 + \rho_o$$

$$= \begin{cases} \rho_1 + \sigma_{m^*}(S^+_{o_{max}}) & t_i < t_{lim}, \; S^+_{o_{max}} \leq T_{h2} \\ 0 & otherwise \end{cases} \quad (6)$$

.where $t_{lim}$ is the specified time limit

$T_{h2}$, is a threshold level

$\sigma_{m^*}$, represents an optimized version of the hyperbolic tangent activation function adapted from [28].

To assure the update procedure during learning, a Monte Carlo run is performed on this routine twice.

We may decide to stop learning by setting $\rho_1 = 0$, but we believe that learning is a continual process and there is no end in sight except when the system is dead.

### 4.1.5 Inference:

Inference is achieved by first computing a second-order (level-2) mismatch - a prediction on its own activation. This process is described as follows:

$$k_{dev(2)} = |I_s - I_{g(c)_i}|$$

$$\approx n * \int_n k_{dev(2)_{av}}$$

(7)

.where

$I_{g(c)_i}$ = winning integers obtained from (5) and

$I_s$ = a new input sequence at the next time-step

Note that the average mismatch at level-2 is defined as:

$$k_{dev(2)_{av}} = \int_n k_{dev(2)} \qquad (8)$$

.from which we obtain:

$$k_r = \iint_n k_{dev(2)} \qquad (9)$$

.where $k_r$ is a rating factor that can be used to evaluate inference performance over time.

Using (7) and (5), a predictive interpolation is performed as:

$$P_r = C_{ao}(\cdots, i) = \begin{cases} C_{ao} & k_{dev(2)} \equiv \min(k_{dev(2)}),\ i = 1,2,\ldots n,\ n \in Z \\ 0 & otherwise \end{cases}$$

$$.and\ P_r^T = \frac{(P_r + I_s)}{2} \qquad (10)$$

Note that $P_r^T \approx P_r$ for ideal matches.

A memorization process is finally performed to extract all successful integer matches by a superimposition operation on the virtual store as:

$$S_{t_{h(}}(\cdots, i) = P_r$$

$$\begin{cases} i = 1,2,\ldots,n \\ i, n \in Z \end{cases} \qquad (11)$$

Note that the prediction $P_r$ is updated by a counter defined by $i$ at each time step.

**4.2 Post-Prediction Phase**

In post-predictive phase, we perform a third and final-order mismatch operation (level-3) conditioned on a new permanence threshold, say, $\rho_2$ as:

$$k_{dev(3)} = |I_{S_{current}} - S_{t_h}(\cdots, j)|$$
$$= \begin{cases} 1 & \rho_2 \leq \rho_{2_{lim}}, \; j \in Z, 0 \leq \rho_{2_{lim}} \leq 1 \\ 0 & otherwise \end{cases} \quad (12)$$

We also compute a row-wise overlap as:

$$S^+_{k_{(t)}} = \sum_{\rho_2 \leq 1}^{(\cdots)} k_{dev(3)}$$

$$(13)$$

Finally, we extract the possible match from the memory store by a conditional selection procedure using a prefix search [29]:

$$P_{r(post)} = S_{th(i_p)}, \; \forall j$$
$$= \begin{cases} i_p, & S^+_k \geq T_{h3}, \quad i_p \subseteq j, \; T_{h3} \in Z \\ \varnothing & otherwise \end{cases} \quad (14)$$

$T_{h3}$, is computed as:

$$T_{h3} = lo/2 \quad (15)$$

.where $lo$ is the length of novel deviants.

In practical terms, this is typically a backward pass for the exemplar at the previous time-step.

**4.3 Back-ward Additive Deviant Computing**

The DLA's predictive capability may be extended by performing a deviant feed-forward extrapolation using a novel technique we refer to here as "backward additive deviant computing (BADC)". BADC is computed as follows:

1. Train a temporal DLA state observer with a temporal sequence of exemplars using the values obtained in (11). This generates a memory of deviants
2. Using the nth prediction in (11) as a deviant and the previous (n-1) deviants as standards perform an aggregated deviant operation as:

$$K_{avg}^t = \frac{\sum_{j=1}^{n} \left| K_n^t - K_{seq}^t \right|}{n} \quad \{K_{seq}^t = K_j^t, \quad j = 1,2,..n-1, j \in Z^+ \tag{16}$$

Where,

$K_n^t$ = nth memorized sequence chunk at time, t

$K_{seq}^t$ = memorized sequence chunks at time steps of t

$n$ = number of previously memorized sequence chunks

3. Using (11), compute the deviants numeric prediction as:

$$K_p^t = K_{avg}^t ,+ K_n^t \tag{17}$$

The prediction ($K_p^t$) formed in (17) can then be combined with that in (15) to create a memory field effect (MFE).

## 5. Experiments and Results

### 5.1 Experiment 1: Classification Tasks

Experiments have been performed using three benchmark datasets for the classification and similarity estimation problems. The first two datasets are specifically a classification problem; these are the IRIS dataset which is a plant species dataset [30] for categorizing the Iris plant into three different classes, the HEART dataset [31] for categorizing a heart condition as either an absence or a presence of a heart disease and the Word-Similarity dataset [32] for estimating similarity scores.

We begin by defining some initial experimental parameters for the HTM-CLA and the DLA. For the HTM-CLA, we build a hierarchy of cells using a Monte-Carlo simulation constrained by their corresponding permanence values. This modification is less computationally intensive than the technique of combinatorics proposed in [12]. The parameters for the simulation experiments are given in Appendix B. The classification accuracies for HTM-CLA and DLA are presented in Tables 1 and 2 respectively. We use a classification metric termed the mean absolute percentage classification accuracy (MAPCA). MAPCA is similar to the mean absolute percentage error (MAPE) used in [33]. MAPCA is computed as:

$$MAPCA = \left( \frac{\sum_{i}^{n}(|y_{(i)} - \hat{y}_{(i)}| < tol)}{n_z} \right) * 100 \qquad (18)$$

.where $y$ = the observed data exemplars

$\hat{y}$ = the model's predictions of $y$

$n_z$ = size of the observation matrix, and

$tol$ = a tolerance constraint.

TABLE 1: Classification accuracies for the HTM-CLA

|  | IRIS Dataset | HEART Dataset | Word-Similarity Dataset |
|---|---|---|---|
| **Percent accuracy (%)** | 77.03 | 75.07 | 79.35 |

TABLE 2: Classification accuracies for the DLA

|  | IRIS Dataset | HEART Dataset | Word-Similarity Dataset |
|---|---|---|---|
| **Percent accuracy (%)** | 86 | 70 | 72.5 |

**5.2 Experiment 2: Variational Learning Extent Study**

In this experiment, we study the DLAs predictions for different learning extents. For each run, we increment the learning extent in 5 steps of width equal to 50, and with a starting value of 50. The variational response of the DLA for the IRIS dataset is shown in Figure 2.

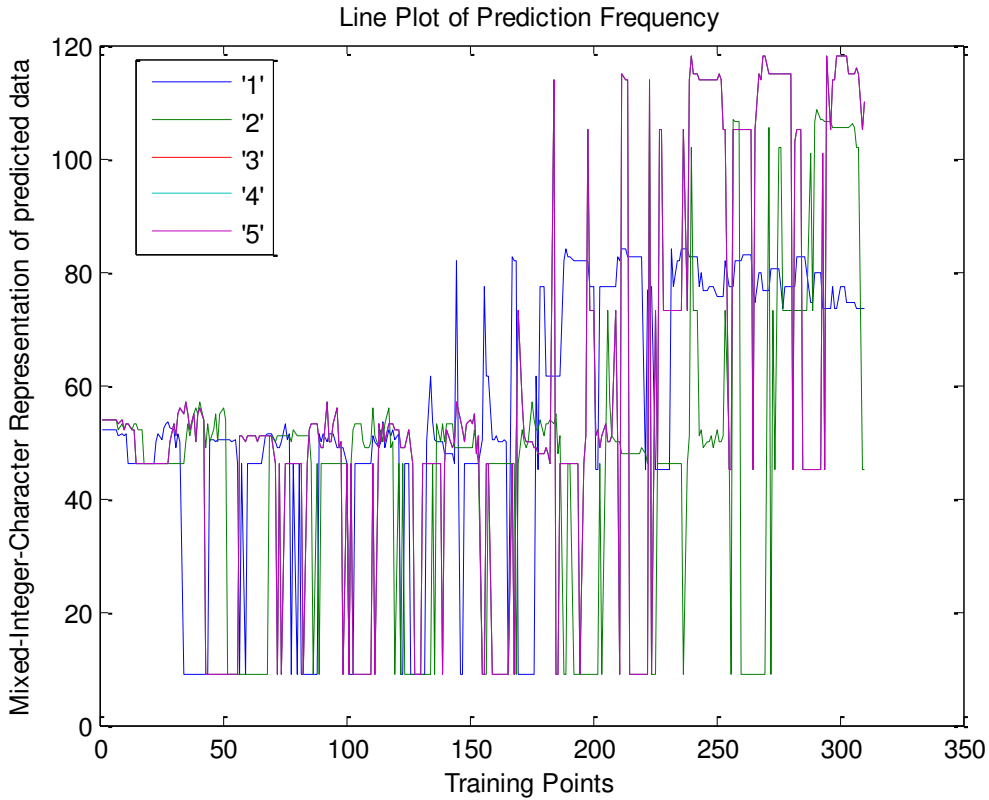

FIGURE 2: Variational response of the DLA for the IRIS dataset; the heat plots reprsesnts a learning extent of 50 (1) to 250 (5).

## 6. Discussions

The results show a competitive performance for both the HTM-CLA and DLA using the specified parameters – see Appendix B for the classification task. While the DLA outperformed the HTM-CLA on the IRIS dataset test, the HTM-CLA outperformed the DLA on the HEART and Word-Similarity datasets. This may be attributed to variations in parameters and HTM-CLA's highly sparse representation. In particular, the effect of learning extent show how the DLA can learn to remember and forget without the use of any special gating mechanisms i.e. the DLA's memory is a function of the number and strength of its learning integers which may said to be equivalent to the number of learning synapses in a HTM model. The learning strength (see Figure 2) is increased with higher learning extent (remembering) and decreased with lower learning extent (forgetting). The use of long chain of integers is thus in accordance with large number of synapses as defined in [16].

## 7. Conclusion and Future Work

A novel algorithm for cortical learning based on the MMN effect has been developed. Our model can learn long range context, classify and predict forward in time. With the DLA, accuracies as high as 95% is attainable and realistic performances can be synthesized by the algorithm, provided that the learning extent is appropriately tuned and that the data or the observation process is repeatable. It is possible to implement forgetting and remembering by dynamically adjusting the learning extent. Further details of this effect and its use in a multiple of different tasks such as language modelling and sequence learning including real-time embedded systems will be examined in a future version of this paper.

## 8. Original Contributions to Knowledge

In this study, a novel theoretical framework/algorithm for predictive solutions, the Deviant Learning Algorithm (DLA), has been developed and presented in this research paper. The DLA learns a sparse mixed-integer representation of an input sequence in its receptive field and also is the first machine learning algorithm that used the idea of the MMN effect on some historical and modern state of the art real world benchmark datasets. The DLA is targeted at machine learning tasks where the need to solve in a predictive manner, computational and day-to-tasks is a priority.

In particular, this work hypothesizes that the brain learns a sparse statistical representation of its world using a temporally adapted deviant input and a long list of standards; an approach which is based on the MMN effect. Our implementation algorithm mimics this effect using a concept referred to as the Real Absolute Deviation (RAD). With the RAD, it is possible to temporally generate a sparse mixed-integer representation of the input world using a long sequence of standard integers. We also introduced in this research a concept referred to as backward additive deviant computing;

this permits interpolation/extrapolations to be made on the recognition units of a standard DLA.

While, this research proposes some novel ideas, we do not specifically claim not to use some useful ideas from other researchers. For instance, the concept of overlap and sparse coding have been utilized in a somewhat similar way to that of Ahmad and Hawkins [12]. However, instead of logical sum of products used in computing the overlap, we apply the logical sum of real absolute deviations of standards from a stimulating deviant input signal in a bid to disentangle the underlying causes of temporal patterns in the given observation. This approach, allows us to apply the principles of statistical processing to any class of problem – both real and synthetic.

Our goal, therefore, is not to replicate in its entirety, the MMN effect in software, but to mimic a subset of this phenomena for general problem solving tasks which are difficult for even the most powerful high-end computers, but relatively easy for humans to solve.

**Conflicting Interests**

The authors declare that they have no conflict of interests regarding the development and publication of this paper.

**Acknowledgement**

This study received no funding from any source or agency. Source codes for the HTM-CLA and DLA simulations are available at the Matlab central website:

www.matlabcentral.com

**APPENDIX A: DLA Pseudocode**

**Initialize** time counter, time limits ($Counter_n$), memory store, and permanence threshold

`Counter=1`

**Generate** a long list of Integers (standards), $I_{g_{(1)}}$, $i=1...n, \ n \in Z$

**FOR** each exemplar

    Evaluate 1$^{st}$ order Mismatch: $k_{dev(1)}$    // Pre-prediction Phase

      Compute Deviant-Overlap

      Select Winner Integers

      Store Winner Integers

      Update permanence threshold   // Learning

    Evaluate 2$^{nd}$ order Mismatch: $k_{dev(2)}$

      Perform a predictive interpolation in time, t

**If** $k_{dev(2)}$ is less than min ($k_{dev(2)}$)

    Extract the causal Integers    // Integers maximally responsible for the memorization process

    Evaluate 3$^{rd}$ order Mismatch: $k_{dev(3)}$    // Post-predictive phase

      Compute Deviant-Overlap

      Extract novel memories from store

`Counter=Counter+1`

**Until** Counter = $Counter_n$

## APPENDIX B: Experimental Parameters

HTM-CLA uses a different architecture from the DLA with different set of parameters. HTM-CLA parameters for the IRIS, HEART and Word Similarity Datasets is provided in Table B.1 while that of the DLA is given in Table B.2

TABLE B1: HTM-CLA parameters

| Parameter | IRIS Dataset | Heart Dataset | Word-Similarity Dataset |
|---|---|---|---|
| Desired local activity | 3 | 3 | 3 |
| Minimum overlap | 90 | 210 | 123 |
| Initial permanence value | 0.4 | 0.4 | 0.4 |
| Number of Monte-Carlo Runs | 1000 | 1000 | 1000 |
| Tolerance constraint | 0.05 | 0.05 | 0.05 |

TABLE B2: DLA parameters

| Parameter | IRIS Dataset | Heart Dataset | Word-Similarity Dataset |
| --- | --- | --- | --- |
| Learning extent | 200 | 200 | 200 |
| Time limit | 70 | 70 | 70 |
| Initial permanence value | 0 | 0 | 0 |
| Store threshold | 120 | 120 | 120 |
| Tolerance constraint | 0.05 | 0.05 | 0.05 |